\documentclass[10pt,twocolumn,letterpaper]{article}
\usepackage{graphicx}
\usepackage{multirow}
\usepackage{array}
\usepackage{booktabs}
\usepackage{caption}
\usepackage{cvpr}              % To produce the CAMERA-READY version
\definecolor{cvprblue}{rgb}{0.21,0.49,0.74}
\usepackage[pagebackref,breaklinks,colorlinks,allcolors=cvprblue]{hyperref}

\def\paperID{*****} % *** Enter the Paper ID here
\def\confName{CVPR}
\def\confYear{2025}

\title{EVQAScore: A Fine-grained Metric for Video Question Answering Data Quality Evaluation}

\author{
Hao Liang\thanks{Equal contribution.}$^*$ \\
Peking University\\
{\tt\small hao.liang@stu.pku.edu.cn}
\and
Zirong Chen\footnotemark[1]$^*$\\
Beijing Institute of Technology\\
{\tt\small 1120223580@bit.edu.cn}
\and
Hejun Dong\footnotemark[1]$^*$\\
Beihang University\\
{\tt\small 22373333@buaa.edu.cn}
\and
Wentao Zhang\\
Peking University\\
{\tt\small wentao.zhang@pku.edu.cn}
}

\begin{document}
\maketitle
\begin{abstract}
Video question-answering (QA) and video-caption data play a central role in video understanding. Evaluating the quality of both VideoQA and video-caption data for training video large language models (VideoLLMs) presents a significant challenge. While several methods have been proposed for assessing video-caption quality, there is a lack of dedicated evaluation methods specifically for VideoQA. To address this gap, we introduce EVQAScore, a reference-free method that employs patch retrieval for fine-grained evaluation of video-caption and VideoQA data quality. Additionally, we utilize a keyword extraction method to evaluate longer captions (previous methods like PAC-S and EMScore are limited by CLIP’s 77-word constraint) and to improve semantic understanding. Furthermore, we incorporate frame sampling to enable faster evaluation without sacrificing performance. Our approach achieves state-of-the-art (SOTA) performance on the VATEX-EVAL benchmark for video-caption evaluation, with scores of 28.9 for Kendall correlation and 37.4 for Spearman correlation, surpassing the previous method PAC-S++ by 0.8 and 1.0 points, respectively. By using EVQAScore for data selection, we achieve SOTA results with only 12.5\% of the original data volume, outperforming PAC-S, which required 100\% of the data. Our code is available at \url{https://anonymous.4open.science/r/EVQAScore-FFA6/}.
\end{abstract}

\section{Introduction}
\begin{figure}[h] % 'r' 表示图像靠右，'0.5\textwidth' 是图像宽度
  %\vspace{3mm}  % 控制图片和上方文字的距离
  \centering
  \includegraphics[width=0.49\textwidth]{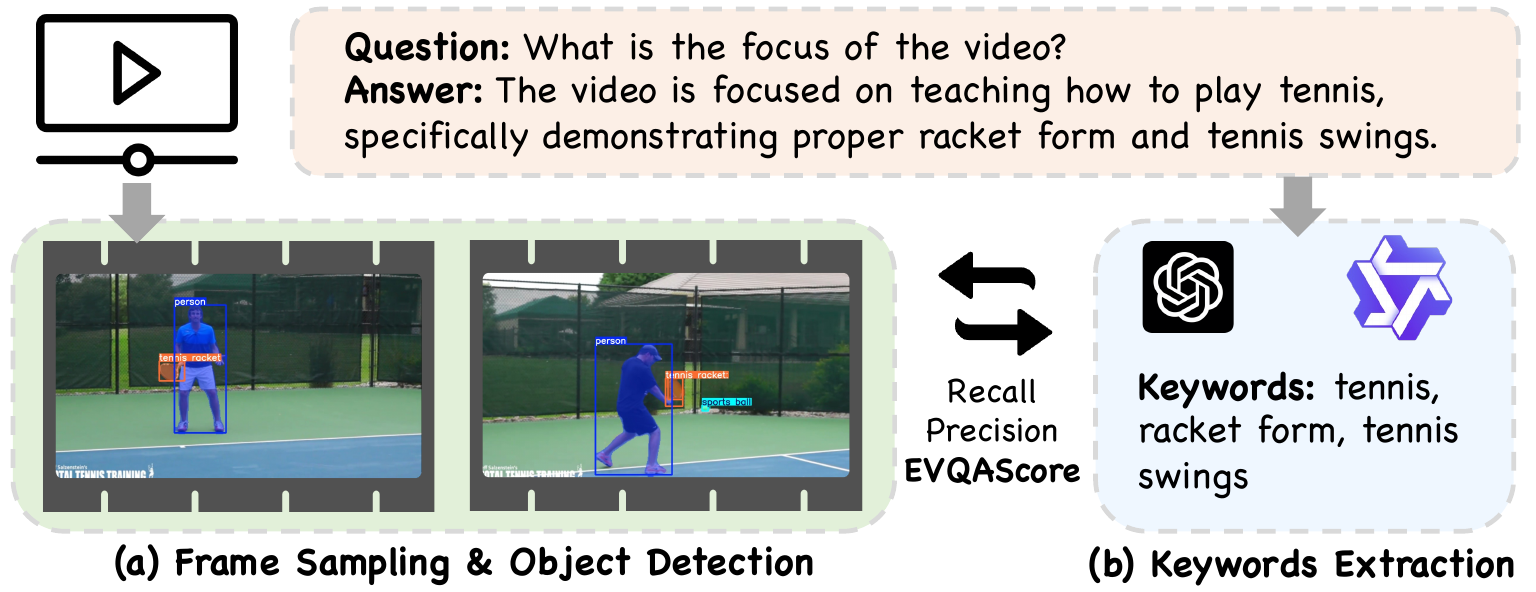}
  \caption{We provide an example of the EVQAScore evaluation process. In (a), we sample frames and apply YOLO for patch recognition. In (b), we extract keywords from the question-answer pairs. Finally, we calculate the EVQAScore based on the precision and recall between patches and keywords.}
  \label{fig:Face_1}
  %\vspace{-4mm}  % 控制图片和下方文字的距离
\end{figure}
With the rapid advancements in large language models (LLMs)~\cite{chatgpt, llama} and multimodal large language models (MLLMs)~\cite{zhao2023survey,wu2023multimodal}, VideoLLMs have achieve competitive performance in traditional multimodal tasks such as visual recognition~\cite{zhang2024vision}, video understanding~\cite{xu2021vlm, tang2023video}, and action recognition~\cite{internvideo2}. Moreover, their excellent language understanding capabilities enable strong performance in text-rich tasks, such as video question-answering~\cite{hu2024bliva} and video-centric dialogues~\cite{internvideo2}.
%prompt engineering has emerged as a technique to enhance model performance~\cite{kojima2022large, wei2022chain, yao2024tree, besta2024graph, yang2024buffer, wang2023plan}, guiding these powerful models to generate more accurate, contextually appropriate, and relevant outputs. By strategically crafting prompts, researchers and practitioners can better leverage the full potential of LLMs and MLLMs across a wide range of tasks, from natural language understanding to complex, multimodal reasoning.

% Among MLLMs, VideoLLMs achieve competitive performance in traditional multimodal tasks such as visual recognition~\cite{zhang2024vision}, video understanding~\cite{xu2021vlm, tang2023video}, and action recognition~\cite{internvideo2}. Moreover, their excellent language understanding capabilities enable strong performance in text-rich tasks, such as video question-answering~\cite{hu2024bliva} and video-centric dialogues~\cite{internvideo2}.
Most existing VideoLLMs focus on modifying model architectures to effectively leverage information from multiple modalities~\cite{internvideo, internvideo2, video-chatgpt, video-llama, videochat}. While model effectiveness is critical, data quality also plays a pivotal role in the success of VideoLLMs. For example, \citet{videochat2, internvideo2} demonstrate that higher-quality training data significantly improves the performance of VideoLLMs. Therefore, evaluating the quality of VideoQA data is essential. On the other hand, previous methods, such as EMScore and PAC-S~\cite{shi2022emscore, sarto2023positive}, have focused on evaluating human-preference video-captions. In this work, we aim to develop an evaluation metric that not only excels in human-preference evaluation but also provides high-quality training data for VideoLLMs. We set the following three tasks as our goals for improvement.
\paragraph{Detailed Information from Video Frames} Existing methods (PAC-S and EMScore) typically use a single frame as the minimum unit for video feature extraction, potentially overlooking the rich information within a single frame. To address this limitation, we employ YOLO~\cite{yolo} to extract key objects from video frames, enabling the utilization of diverse information within a single frame.
\paragraph{Better Semantic Understanding} Methods such as PAC-Score~\cite{sarto2023positive} and EMScore~\cite{shi2022emscore} rely on TF-IDF for semantic understanding, which is limited as it only computes word frequency and fails to capture contextual semantics. To address this limitation, we employ LLMs to extract keywords, enabling a more accurate understanding of both QA and caption semantics. Furthermore, in the context of long videos with extended captions and QAs, which often exceed 77 words (with over 30\% of the VCG+~\cite{Maaz2024VideoGPT+} Train Dataset containing examples longer than 77 words), PAC-Score and EMScore become less effective. As a result, there is a need for an improved evaluation method capable of supporting longer videos and arbitrarily lengthy captions and QAs, providing better semantic comprehension.
\paragraph{VideoQA Data Quality Evaluation} As VideoQA becomes increasingly important in areas such as SFT, evaluating the quality of VideoQA data is crucial. Therefore, we need a evaluation method supporting both video-caption and VideoQA quality evaluation.

To improvement the from the three aspects, we introduce EVQAScore, which utilizes LLMs to extract keywords for enhanced caption and QA data understanding. Moreover, we use YOLO~\cite{yolo} to extract fine-grained feature. To support the evaluation of longer videos, we need to control the computational cost. Therefore we efficiently compute EVQAScore through uniform frame sampling, which reduces computational costs by a factor of 30 without affecting results. Compared to previous methods such as EMScore~\cite{shi2022emscore} and PAC-S~\cite{sarto2023positive}, our approach can be applied to videoQA evaluation and achieves superior visual understanding through keyword extraction, outperforming the TF-IDF technique used by EMScore and PAC-S. Additionally, due to its efficiency and keyword extraction capabilities, our method can handle extremely long captions and QAs for videos, a feat not possible for EMScore and PAC-S because of the 77-word limit imposed by CLIP. 

The \textbf{core contributions} of this paper are summarized as follows:
\begin{itemize}
    \item  \textbf{New Method:} We introduce a novel approach, EVQAScore, designed for the evaluation of VideoQA and video-caption data quality. To the best of our knowledge, this is the first systematic effort to evaluate the quality of VideoQA data.
    Our approach leverages LLMs and keyword extraction techniques to improve the understanding of QA data in VideoQA evaluation. Additionally, we employ uniform frame sampling to enhance efficiency and scalability, allowing EVQAScore to be applied to extremely long videos and large datasets.
    % \item \textbf{Efficient VideoQA Data Evaluation:} We use uniform frame extraction, resulting in a computational cost reduction of over 30 times compared to processing all frames. Furthermore, we demonstrate that this reduction does not lead to any performance degradation.
    \item \textbf{SOTA Performance for Video-Caption Evaluation:} We utilize LLM for keyword extraction for better caption and QA understanding. We evaluated the performance of our method on the VATEX-Eval benchmark for video-caption data, achieving a Kendall correlation of 28.9 and a Spearman correlation of 37.4, which are 0.8 and 1.0 points higher than the previous method PAC-S++.
    \item \textbf{SOTA Performance for Filtering VideoLLMs Training Data:} We constructed a dataset comprising 400K video entries, consisting of 200K noisy samples and 200K high-quality samples. Using our EVQAScore and the previous SOTA model PAC-S, we filtered the data, retrieving a significantly lower amount of noisy data than PAC-S. As shown in Figure \ref{fig:Face_1}, our method is the only method that can identify low-quality VideoQA data. Additionally, when evaluated on ActivityNet, MSRVTT, MSVD, TGIF, MVBench, and two longer benchmarks—VideoChatGPT Bench and VideoChatGPT Bench Diverse—our EVQAScore outperformed PAC-S and 100\% of data with only 12.5\% of data.
\end{itemize}

\section{Related Work}
\subsection{Video-Caption Data Quality Evaluation}
% %czr 
\paragraph{Video Data Quality Evaluation}
Traditional image and video-captioning evaluations have relied on metrics like BLEU\cite{papineni2002bleu}, METEOR\cite{banerjee2005meteor}, ROUGE\cite{lin2004rouge}, CIDEr\cite{vedantam2015cider}, and SPICE\cite{anderson2016spice}. These metrics, based on n-gram matching, primarily focus on lexical similarity between generated and reference captions. While effective in text-based tasks like machine translation, they struggle to capture the semantic richness of video content, leading to poor correlation with human judgments, especially when evaluating dynamic, time-sensitive visual elements in videos.
\paragraph{Video-Caption Data Quality Evaluation}
Recently, methods like CLIP-Score\cite{hessel2021clipscore} leverage pre-trained vision-language models to measure the coarse-grained alignment between video frames and captions. However, CLIP-Score lacks fine-grained analysis. To address this, Emscore\cite{shi2022emscore} introduced fine-grained matching between video frames and specific words in the captions, significantly improving semantic alignment. Building upon Emscore, PAC-Score\cite{sarto2023positive} further enhances the precision of video-text alignment by incorporating a positive-augmented CLIP model, leading to a more comprehensive evaluation of video content.

\subsection{Video Large Language Models.}
Recently, inspired by the remarkable understanding capabilities of LLMs~\cite{add:g-refine}, multimodality breakthroughs~\cite{add:g-refine, add:q-refine, add:agiqa-3k, add:ntire2024, li2024cmcbench, li2024misc, luo2024llm, lin2024draw, huang2024can}, and pre-trained models, researchers have started using LLMs to understand videos, achieving SoTA results~\cite{videochat2,video-llama,minigpt4,internvideo2,video-llava}. VideoLLaMA~\cite{video-llama} is one of the pioneering studies in VideoLLMs, utilizing a visual encoder and a video Q-Former projector to understand videos. However, due to its Q-Former structure, the computational cost is high. To address this, subsequent works such as VideoLLaVA~\cite{video-llava}, LLaVA-Next~\cite{llava} adopted the LLaVA~\cite{llava, llava1.5} MLP structure, significantly reducing computational costs while still achieving SoTA performance. Similarly, MiniGPT4Video~\cite{minigpt4} uses an MLP adapter for efficient training.

Another notable series of models includes VideoChat, VideoChat2, InternVideo, and InternVideo2~\cite{videochat, videochat2, internvideo, internvideo2}. These models utilize an enormous amount of data to train a transformer-structured adapter, achieving SoTA performance. By leveraging large-scale datasets and advanced transformer architectures, these models excel in comprehending and processing multimodal video content, further pushing the boundaries of video understanding capabilities.
%写一列的一半左右

\section{Method}
\begin{figure*}[ht]
  \includegraphics[width=\textwidth]{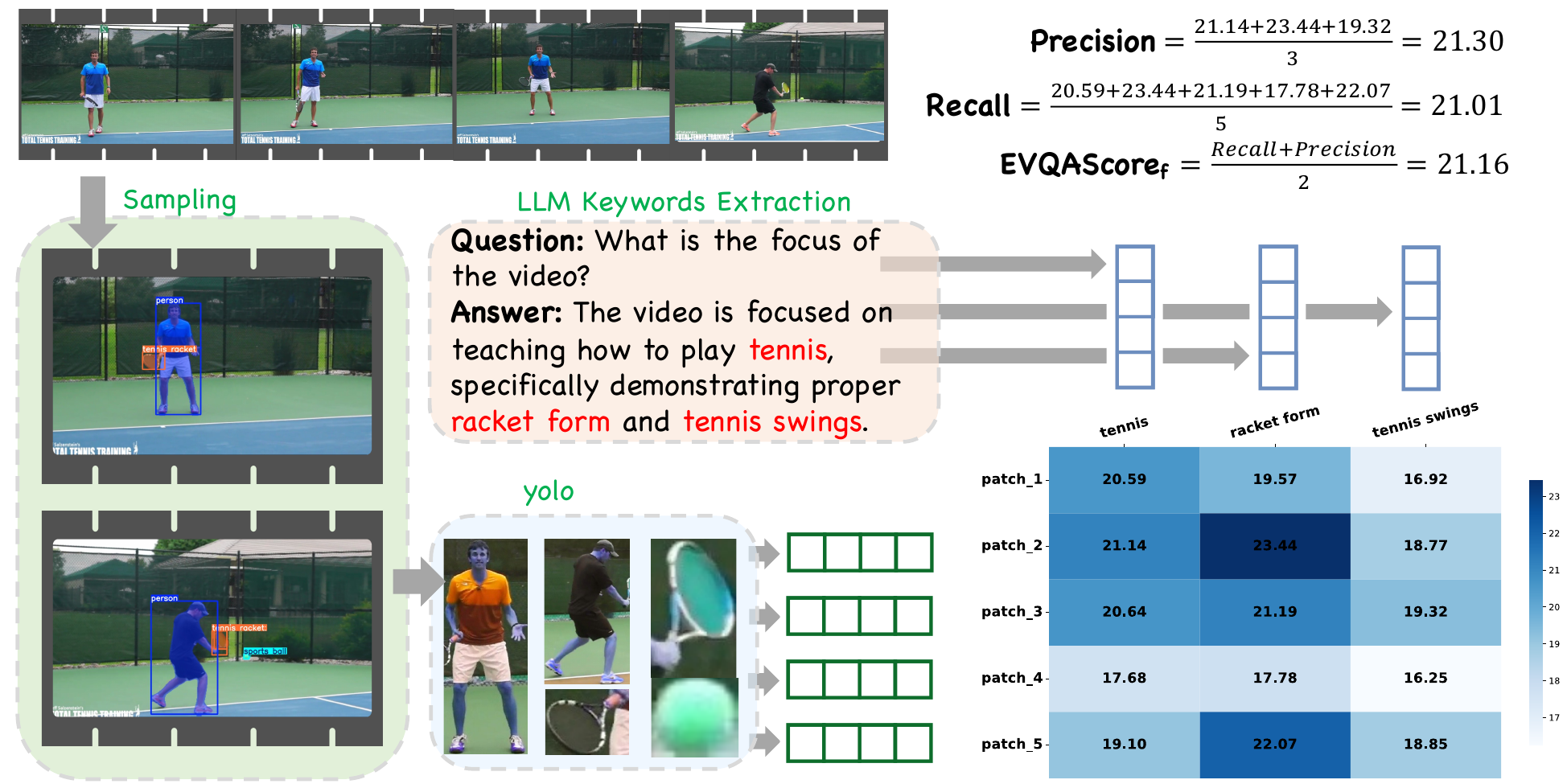}
  \caption{We illustrate the computation of $\text{EVQAScore}_f$. First, fine-grained patches are extracted from the sampled frames. Next, keyword extraction is performed to enhance caption and question-answer semantic understanding. Finally, the patches and keywords are combined to calculate precision and recall, resulting in the EVQAScore.}
  \label{fig:Main_Algorithm}
  %\vspace{-4mm}
\end{figure*}
% \subsection{Overview of PAC-Score}
% PAC-Score provides a comprehensive evaluation of video-text alignment by combining both coarse-grained and fine-grained matching techniques. The coarse-grained embedding matching score between the video $V$ and the generated caption $X$ is defined as follows:

% \begin{equation}
% PACScore(X, V)_c = f_X^T f_V
% \end{equation}

% where $f_V$ represents the video frame embedding and $f_X$ represents the sentence token embedding. For a more refined matching, we apply fine-grained embedding matching to achieve frame-token alignment. The precision ($P$) and recall ($R$) for fine-grained matching are calculated as follows:

% % \begin{equation}
% % P(X, V)_f = \frac{1}{|X|} \sum_{x_i \in X} \max_{v_j \in V} f_{x_i}^T f_{v_j}
% % \end{equation}

% % \begin{equation}
% % R(X, V)_f = \frac{1}{|V|} \sum_{v_j \in V} \max_{x_i \in X} f_{x_i}^T f_{v_j}
% % \end{equation}

% The fine-grained embedding matching score is then computed using the F1-score formula:

% \begin{equation}
% \text{EVQAScore}(X, V)_f = \frac{2PR}{P + R}
% \end{equation}

% Finally, the overall PAC-Score is the average of the coarse-grained and fine-grained scores:

% \begin{equation}
% \small
% \text{EVQAScore}(X, V) =\frac{\text{EVQAScore}(X, V)_c + \text{EVQAScore}(X, V)_f}{2} 
% \end{equation}

\subsection{Problem Formulation}
Inspired by EMScore and PAC-S, we extend these video-caption data quality evaluation methods to assess both video-caption and VideoQA data quality with a more fine-grained metric, EVQAScore. We now present the problem formulation:

Given a video \(V\) and a corresponding question-answer pair \((Q, A)\), our goal is to evaluate the quality of the VideoQA data using the proposed EVQAScore metric. This metric evaluates the data's suitability for training VideoLLMs by scoring based on several criteria, including the semantic relevance and informativeness of the answer \(A\) in relation to the video \(V\) and the question \(Q\).

Formally, for a given data instance \((V, Q, A)\), where \(V=\{v_i\}_{i=1}^{|V|}\) (\(|V|\) is the number of video frames), we compute the EVQAScore to assess the quality of \((V, Q, A)\):
\[
\text{EVQAScore}(V, Q, A)
\]
Our objective is to filter and select high-quality data based on the EVQAScore to 1) select data aligned with human preferences, and 2) improve the performance and training efficiency of VideoLLMs.

\subsection{Data Preparation for EVQAScore}
To address the challenge of evaluating long videos efficiently, we adopt a uniform frame sampling strategy to reduce the number of frames processed while retaining critical information.
\paragraph{Frame Sampling}
By sampling frames at regular intervals, we capture essential content and minimize computational costs. The frame sampling process is defined as:
\begin{equation}
S = \{v_i \mid i = k \cdot l + 1, \; k \in \mathbb{Z}_{\geq 0}, \; k \cdot l \leq m \}
\end{equation}
where \(S\) represents the set of sampled frames, \(v_i\) denotes the \(i\)-th frame in the video $V$, \(m\) is the total number of frames, and \(l\) is the sampling interval. This approach reduces redundancy while ensuring a diverse representation of sampled frames. 

From Table \ref{tab:keyframe_extraction}, it is evident that when the sampling interval \(l\) is set to 10, 20, 30, 40, 50, or 60, the EVQAScore remains stable, consistently outperforming the EMScore, PAC-S, and PAC-S++ metrics. This suggests that frame sampling does not significantly influence the EVQAScore calculation. In our experiment, we select a sampling interval of \(l = 30\), which reduces computational costs without any degradation in performance.

After sampling the frames, we extract fine-grained patch features for more detailed matching. Additionally, we perform keyword extraction to enhance semantic understanding.
\paragraph{Key Object Extraction}
For each frame \(v_i\) in \(S\), we use YOLO to extract key objects $O_i$, saying:
\begin{equation}
O_i=\text{YOLO}(v_i)
\end{equation}
where \(O_i = \{o_{i,j}\}_{j=1}^{|O_i|}\) stands for a list of object regions in a frame. Each object \(o_{i,j}\) is a region in the original video frame \(v_i\) containing meaningful items. Then we concat all \(O_i\) to get the final object sequence:
\begin{equation}
O=\text{concat}_{i=1}^{|S|}(O_i)
\end{equation}
\begin{figure}[ht]
  \includegraphics[width=0.47\textwidth]{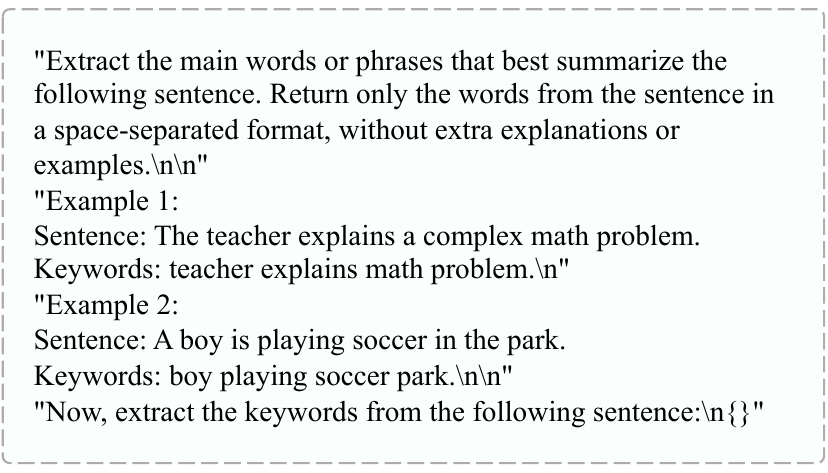}
  \caption{Prompt used for keywords extraction.}
  \label{fig:Keywords}
\end{figure}
\paragraph{Keywords Extraction}
Given a text input \( X \), we use a large language model (LLM) to extract a list of key phrases $K = \{k_1, k_2, \ldots, k_n\}$ as follows:
\[
\{k_1, k_2, \ldots, k_n\} = \text{LLM}_{\text{extract}}(X, P)
\]
% \[
% K = \{k_1 \oplus k_2 \ldots \oplus k_n\}
% \]
where \( \text{LLM}_{\text{extract}} \) represents the extraction function of the language model, and \( P \) denotes the prompt designed to guide the extraction of relevant phrases from the input \( X \). This approach improves the understanding of caption semantics and is particularly effective for question answering, especially when the caption is lengthy.
\subsection{EVQAScore for Video-Caption Data Quality Evaluation}
The improved scoring method now combines the coarse-grained score and the modified fine-grained score, which includes fine-grained feature and keyword-based matching. This enhanced PAC-Score improves efficiency when handling long videos and complex text descriptions, focusing on the most essential frames and phrases.
%\vspace{-2mm}
\paragraph{Coarse-Grained EVQAScore}
Coarse-grained score of  $\text{EVQAScore}(V, X)_c$ is:
$$
\text{EVQAScore}(V, X)_c=f_V^{\top} f_X
$$
where $f_V$ is the embedding of the video, which is calculated by pooling the feature of video frames in sampled frames $S$, respectively. $f_X$ is the feature of caption or QA pair.
\vspace{-2mm}
\paragraph{Fine-Grained EVQAScore}
First, the precision ($P$) and recall ($R$) for the caption $X$ are updated using their keywords as follows:
$$
\begin{aligned}
& P(V, X)_f = P\left(O, K\right)_f = \frac{1}{|X|} \sum_{k_j \in K} \max_{o_i \in O} f_{o_i}^{\top} f_{k_j} \\
& R(V, X)_f = R\left(O, K\right)_f = \frac{1}{|O|} \sum_{o_i \in O} \max_{k_j \in K} f_{o_i}^{\top} f_{k_j}
\end{aligned}
$$
% $$
% \begin{aligned}
% & P(V, X)_f = \frac{1}{|X|} \sum_{k_j \in K} \max_{o_i \in O} f_{o_i}^{\top} f_{k_j} \\
% & R(V, X)_f = \frac{1}{|V|} \sum_{o_i \in O} \max_{k_j \in K} f_{o_i}^{\top} f_{k_j}
% \end{aligned}
% $$
where $f_{o_i}$ and $f_{k_j}$ represent the CLIP embedding scores of object $o_i$ and keyword $k_j$. Subsequently, the fine-grained EVQAScore is calculated as:
$$
\text{EVQAScore}(V, X)_f = 2 \frac{P(V, X)_f \cdot R(V, X)_f}{P(V, X)_f + R(V, X)_f}
$$
Finally, the overall EVQAScore is computed by combining the original coarse-grained score with the updated fine-grained score, which incorporates both fine-grained and coarse-grained matching.
\begin{equation}
\scriptsize
\text{EVQAScore}(V, X) = \frac{\text{EVQAScore}(V, X)_c + \text{EVQAScore}(V, X)_f}{2}
\end{equation}
This formula ensures that both visual and textual representations are aligned more effectively, improving precision, recall, and computational efficiency by focusing on fine-grained patches and keywords.

\subsection{EVQAScore for VideoQA Data Quality Evaluation}
We compute the EVQAScore for VideoQA data using the following formula:
$$ \text{EVQAScore}(\text{Concat}(Q, A), V) $$
In this formulation, \(\text{Concat}(Q, A)\) represents the concatenation of the question \(Q\) and the answer \(A\). By employing a keyword extraction technique, our EVQAScore is able to accurately capture the semantics of the QA pair.
% $$ \text{EVQAScore}(\text{Concat}(Q, A), V) \cdot \log(n + 1) $$
%while \(n\) denotes the number of extracted keywords. We posit that a higher number of extracted keywords often indicates a richer and more informative caption, thereby warranting a higher score. We show the advancement of our EVQAScore in Section \ref{sec: Experiments}.

\begin{table}[h!]
\caption{We compare the time required to evaluate the VATEX-EVAL video-caption dataset. EVQAScore consistently outperforms the baselines, while maintaining an acceptable computational cost. The evaluation is conducted using 8 NVIDIA H20 GPUs.}
\centering
\resizebox{0.49\textwidth}{!}{
\begin{tabular}{lcccc}
\toprule
\textbf{Method} & \textbf{Time (s)} & Kendall & Spearman\\ 
\midrule
EVQAScore (Interval=60)       & 2283.9   & 28.9 & 37.4       \\ 
EVQAScore (Interval=50)       & 2582.6   & 29.0 & 37.5       \\ 
EVQAScore (Interval=40)       & 3177.0   & 29.0 & 37.6     \\ 
EVQAScore (Interval=30)       & 3986.0   & \textbf{29.1} & \textbf{37.6}       \\ 
EVQAScore (Interval=20)       & 5769.3   & \textbf{29.1} & \textbf{37.6}        \\ 
EVQAScore (Interval=10)       & 9774.6  & \textbf{29.1} & \textbf{37.6}        \\ 
EMScore & 887.4 & 23.2 & 30.3 \\
PAC-S & 927.9 & 25.1 & 32.6 \\
PAC-S++ & 998.6 & 28.1 & 36.6 \\
\bottomrule
\end{tabular}
}
\label{tab:keyframe_extraction}
%\vspace{-4mm}
\end{table}

\begin{table}[h!]
\caption{Performance Comparison on VATEX-EVAL Dataset (No Reference). The table presents the Kendall $\tau_b$ and Spearman $\rho$ values, comparing Emscore, PAC-Score, and our proposed EVQAScore. EVQAScore achieves the best performance.}
\centering
\resizebox{0.49\textwidth}{!}{
\begin{tabular}{lcc}
\toprule
\textbf{Metric} & \textbf{Kendall $\tau_b$} & \textbf{Spearman $\rho$} \\ 
\midrule
\textbf{Emscore}         & 23.2           & 30.3     \\ 
\textbf{PAC-S}       & 25.1           & 32.6     \\ 
%\rowcolor{blue!15} 
% & \textcolor{blue}{(+0.5)} & \textcolor{blue}{(+0.7)} \\
\textbf{PAC-S++}        & 28.1  & 36.4 \\
\textbf{EVQAScore(Alpha-CLIP)}        & 26.6  & 34.5 \\
\textbf{EVQAScore(PAC-S)}        & 26.4  & 34.3 \\
\textbf{EVQAScore(PAC-S++)}        & \textbf{28.9}  & \textbf{37.4} \\
% & \textcolor{blue}{(+4.7)} & \textcolor{blue}{(+5.9)} \\                     
\bottomrule
\end{tabular}
}
\label{tab:vatex_eval_comparison}
%\vspace{-4mm}
\end{table}

%和人类对齐(VATEX-Eval)
%\subsection{Video-Caption Evaluation}
\begin{table*}[h]
    \centering
    \caption{Performance of VideoLLaVA and VideoLLaMA2 on ActivityNet, MSRVTT, MSVD, and TGIF. The results demonstrate that, with only 12.5\% of the data filtered by EVQAScore, the VideoLLMs outperform the baseline using 100\% of the data. Furthermore, our approach outperforms PAC-S in data filtering for both models across all evaluation datasets.}
    \small
    \resizebox{\textwidth}{!}{ %
    \begin{tabular}{lcccccccccc}
        \toprule
        \multirow{2}{*}{\textbf{Method}} & \multirow{2}{*}{\textbf{Model}} & \multirow{2}{*}{\textbf{Data Proportion}} & \multicolumn{2}{c}{\textbf{ActivityNet}} & \multicolumn{2}{c}{\textbf{MSVD}} & \multicolumn{2}{c}{\textbf{MSRVTT}} & \multicolumn{2}{c}{\textbf{TGIF}} \\
        \cmidrule(r){4-5} \cmidrule(r){6-7} \cmidrule(r){8-9} \cmidrule(r){10-11}
        & & & \textbf{Accuracy} & \textbf{Score} & \textbf{Accuracy} & \textbf{Score} & \textbf{Accuracy} & \textbf{Score} & \textbf{Accuracy} & \textbf{Score} \\
        \midrule
        Baseline & \multirow{5}{*}{\textbf{VideoLLaMA2}} & 100\% & 28.56 & 1.36 & 40.04 & 1.99 & 36.37 & 1.80 & 32.13 & 1.59 \\
        PAC-S & & 25\% & 41.17 & 2.07 & 71.46 & 3.53 & 62.17 & 3.08 & 57.92 & 2.87 \\
        PAC-S &  & 12.5\% & 45.41 & 2.28 & 81.31 & 4.04 & 62.91 & 3.12 & 57.00 & 2.82 \\
        EVQAScore &  & 25\% & 56.06 & \textbf{2.78} & \textbf{81.55} & \textbf{4.05} & \textbf{70.23} & \textbf{3.47} & \textbf{59.58} & \textbf{2.95} \\
        EVQAScore  &  & 12.5\% & \textbf{56.15} & \textbf{2.78} & 81.13 & 4.04 & 69.13 & 3.44 & 58.47 & 2.88 \\
        \midrule
        Baseline & \multirow{5}{*}{\textbf{VideoLLaVA}} & 100\% &15.64 & 0.78 & 19.93 & 1.00 & 24.97 & 1.24 & 11.72 & 0.58 \\
        PAC-S &  & 25\% & 32.27 & 1.60 & 46.11 & 2.30 & 50.13 & 2.49 & 41.36 & 2.05 \\
        PAC-S &  & 12.5\% & 35.41 & 1.77 & 52.13 & 2.60 & 53.23 & 2.65 & 43.33 & 2.14 \\
        EVQAScore &  & 25\% & 56.23 & 2.78 & \textbf{81.85} & \textbf{4.07} & \textbf{69.96} & \textbf{3.47} & \textbf{57.93} & 2.84 \\
        EVQAScore  & & 12.5\% & \textbf{57.65} & \textbf{2.90} & 80.79 & 4.02 & 69.29& 3.44 & 57.88 & \textbf{2.85} \\
        \bottomrule
    \end{tabular}
    }
    \label{tab:Activitynet}
    %\vspace{-4mm}
\end{table*}
%\vspace{-2mm}
\section{Experiments}\label{sec: Experiments}  
In this section, we apply our EVQAScore method to both video-caption and videoQA datasets. The goal is to address the following questions to assess the effectiveness of our method:\\
\textbf{Q1}: Does our EVQAScore outperform all baseline methods in video-caption data evaluation?\\
\textbf{Q2}: Does our EVQAScore outperform all baseline methods in selecting VideoQA pairs?\\  
\textbf{Q3}: Does our EVQAScore demonstrate superior performance compared to baseline methods when training VideoLLMs?\\
\textbf{Q4}: Can we visualize the advantages of our EVQAScore relative to previous video-caption evaluation methods?\\ 

\subsection{Experimental Setting}
\paragraph{Datasets.}
For Video-Caption data evaluation, to assess the alignment of our metric with human judgment, we utilized the VATEX-EVAL dataset to evaluate the correlation between our scores and human ratings in video-captioning tasks. The VATEX-EVAL dataset is specifically designed to measure the correlation between automatic evaluation metrics and human judgment. It consists of 3,000 validation videos from the VATEX dataset, with a total of 18,000 candidate captions, six for each video.

For the VideoQA evaluation, we selected 200K samples from two datasets: the VideoInstruct100K~\cite{videochatgpt} dataset and the VCG+~\cite{Maaz2024VideoGPT+} 112K dataset. We then augmented these 200K samples with an additional 200K low-quality (noisy) data entries created by substituting each answer with keywords extracted from the answer sentence.

\paragraph{Evaluation Metrics}
For the VATEX-EVAL dataset, we utilize the Kendall $\tau$ and Spearman correlation coefficients to assess the relationship between our rating scores and human ratings.

For VideoQA ability evaluation, following VideoChatGPT~\cite{video-chatgpt, Maaz2024VideoGPT+}, we evaluated the ActivityNet, MSRVTT, MSVD, and TGIF datasets using LLaMA-3.1-Instruct to compare the model's answers against the ground-truth captions and provide a score from 0 to 5 for each caption. Additionally, we used VideoChatGPT Bench and VideoChatGPT Bench Diverse~\cite{video-chatgpt, Maaz2024VideoGPT+} to evaluate model performance on longer videos. Furthermore, we employed MVBench~\cite{li2024mvbench}, a comprehensive benchmark, to evaluate the performance of our method on VideoLLMs.
\begin{table}[ht]
    \centering
    \caption{We compare the noisy data filtered by PAC-S and our EVQAScore. EVQAScore selects significantly less noisy data compared to the previous state-of-the-art method, PAC-S.}
    \resizebox{0.5\textwidth}{!}{ %
    \begin{tabular}{ccccc}
        \toprule
        Method & noisy & videochatgpt & videochatgpt+ & total \\
        \midrule
        \multicolumn{5}{c}{Extract 25\% data} \\
        \midrule
        PAC-S & 27928 & 40012 & 33764 & 101704 \\
        EVQAScore & \textbf{456} & \textbf{53251} & \textbf{47997} & 101704 \\
        \midrule
        \multicolumn{5}{c}{Extract 12.5\% data} \\
        \midrule
        PAC-S & 10832 & 21095 & 18925 & 50852 \\
        EVQAScore & \textbf{15} & \textbf{24640} & \textbf{26197} & 50852 \\
        \bottomrule
    \end{tabular}
    }
    \label{tab:data_filter}
    %\vspace{-4mm}
\end{table}
\paragraph{Models.} We employed the positive-augmented CLIP visual encoder~\cite{sarto2023positive} to extract embeddings from individual frames, applying average pooling to represent the entire video. Similarly, we used the corresponding textual encoder to generate token embeddings, producing a comprehensive caption representation. Additionally, we utilized the LLaMA3.1-8B-Instruct model to identify key terms from question-answer pairs, highlighting the essential elements of the dialogue.

For VideoLLMs, we selected two previous SOTA model Video-LLaVA~\cite{video-llava} and Video-LLaMA2~\cite{video-llama} to evaluate the quality of our selected data.

\paragraph{Baselines.} We used EMScore, PAC-S, and PAC-S++ as baselines. While all three employ the same computational formula, they differ in the specific CLIP models utilized for their calculations.

\paragraph{Settings.} For LLaMA3.1-8B-Instruct, we applied the official hyperparameters from the repository. For the CLIP model, we used the same CLIP score as EMScore~\cite{shi2022emscore}, PAC-S, and PAC-S++~\cite{sarto2023positive}. All experiments were conducted on a machine equipped with 8 NVIDIA H20 GPUs, a 192-core CPU, and 960GB of memory. For YOLO, we used the yolo11x-seg checkpoint and the default settings for key object extraction.

%多模态大模型的效果(允许QA)

% \subsection{Computational Efficient EVQAScore}
% To address \textbf{Q1}, we conduct uniform frame extraction in EVQAScore. We select the sampling interval to be 10, 20 and 30 frames. The results is presented in Table \ref{tab:keyframe_extraction}, they illustrate that using uniform keyframe extraction substantially reduces the time required for managing video data, with significantly shorter processing times compared to not extracting keyframes. For instance, applying a uniform interval of 30 frames results in a processing time of 464 seconds, compared to 18,504 seconds for no keyframe extraction, representing a drastic time reduction. Importantly, this improvement in efficiency comes with nearly \textbf{no impact on performance}, as evidenced by the Kendall and Spearman correlation scores, which remain virtually unchanged across different intervals. This demonstrates that uniform frame selection is a highly efficient approach, achieving considerable time savings without compromising the evaluation quality.

\subsection{EVQAScore for Video-Caption Evaluation}\label{sec:video-caption-evaluation}
% In this study, we compare three automatic evaluation metrics. Emscore is a reference-free evaluation metric that directly computes the alignment between video and caption representations. PAC-Score builds upon Emscore by utilizing CLIP-based visual representations to improve caption evaluation accuracy. Our proposed method, Emscore+, enhances PAC-Score by introducing keyframe extraction and keyword extraction, which further refines the matching process between individual video frames and corresponding caption elements.

% We conduct the evaluation without any reference captions (No Ref) and utilize two correlation coefficients—Kendall $\tau_b$ and Spearman $\rho$—to measure the alignment between the automatic metric scores and human judgment.
To address \textbf{Q1}, we conducted video-caption evaluation experiments on the VATEX-EVAL dataset. The results are presented in Table~\ref{tab:vatex_eval_comparison}. Our findings demonstrate that EVQAScore consistently outperforms both EMScore and PAC-S in correlating with human judgment. Specifically, EVQAScore achieves a Kendall $\tau_b$ score of 26.4 and a Spearman $\rho$ score of 34.3 in the No Reference setting using the PAC-S CLIP model, representing an improvement of 1.3 points in Kendall and 1.7 points in Spearman over PAC-Score. Additionally, further experiments using PAC-S++ checkpoints resulted in SOTA performance, with a Kendall $\tau_b$ score of 28.9 and a Spearman $\rho$ score of 37.4, marking improvements of 0.8 points and 1.0 points over the PAC-S++ baseline. These results indicate that our proposed method better aligns with human perception of video-caption consistency in the No Reference evaluation setting.
\begin{table*}
    \centering
    \caption{Performance of VideoLLaVA and VideoLLaMA2 on VCGBench-Diverse. Our EVQAScore filtered data achieved SOTA performance training VideoLLMs.}
    \resizebox{\textwidth}{!}{ % 将表格等比例缩放到页面宽度
        {\scriptsize % 缩小表格内容的整体字体
        \begin{tabular}{lccccccccccc}
            \toprule
            \multirow{2}{*}{\textbf{Method}} & \multirow{2}{*}{\textbf{Model}} & \multirow{2}{*}{\textbf{Data Proportion}} & \multicolumn{6}{c}{\textbf{Video Understanding Metrics}} & \multicolumn{3}{c}{\textbf{High-Level Metrics}} \\
            \cmidrule(lr){4-9} \cmidrule(lr){10-12}
            & & & \textbf{CI} & \textbf{DO} & \textbf{CU} & \textbf{TU} & \textbf{CO} & \textbf{Avg.} & \textbf{Caption} & \textbf{Spatial} & \textbf{Reasoning} \\
            \midrule
            Baseline & \multirow{5}{*}{\textbf{VideoLLaMA2}} & 100\% & 0.31 & 0.15 & 0.49 & 0.14 & 0.37 & 0.29 & 0.01 & 0.44 & 0.24 \\
            PAC-S 25\% &  & 25\% & 1.56 & 1.87 & 2.17 & 0.93 & 1.82 & 1.67 & 0.79 & 1.66 & 2.70 \\
            PAC-S 12.5\% &  & 12.5\% & 1.57 & 1.80 & 2.10 & 1.01 & 1.87 & 1.67 & 0.43 & 1.76 & 2.87 \\
            EVQAScore 25\% &  & 25\% & 1.86 & 2.48 & 2.64 & \textbf{1.11} & \textbf{2.38} & 2.09 & \textbf{1.33} & 1.99 & 3.17 \\
            EVQAScore 12.5\% & & 12.5\% & \textbf{1.90} & \textbf{2.61} & \textbf{2.70} & 1.08 & 2.32 & \textbf{2.12} & 1.31 & \textbf{2.04} & \textbf{3.41} \\
            \midrule
            Baseline & \multirow{5}{*}{\textbf{VideoLLaVA}} & 100\% & 0.33 & 0.16 & 0.52 & 0.13 & 0.36 & 0.3 & 0.01 & 0.45 & 0.28 \\
            PAC-S 25\% &  & 25\% & 1.63 & 1.92 & 2.20 & 0.87 & 1.79 & 1.68 & 0.75 & 1.78 & 2.81 \\
            PAC-S 12.5\% &  & 12.5\% & 1.62 & 1.86 & 2.15 & \textbf{0.93} & 1.80 & 1.67 & 0.56 & 1.79 & 2.87 \\
            EVQAScore 25\% &  & 25\% & 1.79 & 2.34 & 2.48 & 0.90 & 2.02 & 1.91 & \textbf{1.06} & 2.00 & 3.14 \\
            EVQAScore 12.5\% & & 12.5\% & \textbf{1.80} & \textbf{2.53} & \textbf{2.55} & 0.83 & \textbf{2.05} & \textbf{1.95} & 1.03 & \textbf{2.08} & \textbf{3.39} \\
            \bottomrule
        \end{tabular}
        }
    }
    \label{tab:vcgbench_diverse}
    %\vspace{-2mm}
\end{table*}

\begin{table*}[ht!]
    \centering
    \caption{Performance of VideoLLaVA and VideoLLaMA2 on VCGBench and MVBench. Our EVQAScore filtered data achieved SOTA performance training VideoLLMs.}
    \resizebox{0.8\textwidth}{!}{ % Adjust table width
        \begin{tabular}{lccccccccc}
            \toprule
            \multirow{2}{*}{\textbf{Method}} & \multirow{2}{*}{\textbf{Model}} & \multirow{2}{*}{\textbf{Data Proportion}} & \multicolumn{6}{c}{\textbf{VCGBench}} & \textbf{MVBench}\\
            \cmidrule(lr){4-9} \cmidrule(lr){10-10}
             & &  & \textbf{CI} & \textbf{DO} & \textbf{CU} & \textbf{TU} & \textbf{CO} & \textbf{Avg.} & \textbf{Avg.}\\
            \midrule
            Baseline & \multirow{5}{*}{\textbf{VideoLLaMA2}} & 100\% & 0.60& 0.29 & 0.88 & 0.21 & 1.05 & 0.61 & 34.09\\
            PAC-S& & 25\% & 2.02 & 1.70 & 2.35 & 1.59 & 2.48 & 2.03 & 31.43\\
            PAC-S& & 12.5\% & 2.39 & 2.15 & 2.75 & 1.73 & 2.75 & 2.35 & 34.24\\
            EVQAScore& & 25\% & \textbf{3.06} & 3.09 & \textbf{3.45} & 2.30 & 3.11 & 3.00 & 33.34\\
            EVQAScore& & 12.5\% & 3.03 & \textbf{3.31} & 3.41 & \textbf{2.36} & \textbf{3.17} & \textbf{3.06} & \textbf{36.65}\\
            \midrule
            Baseline & \multirow{5}{*}{\textbf{VideoLLaVA}} & 100\% & 0.59 & 0.30 & 0.87 & 0.20 & 0.96 & 0.58 & 30.68\\
            PAC-S& & 25\% & 2.18 & 1.96 & 2.53 & 1.60 & 2.54 & 2.16 & 37.10\\
            PAC-S& & 12.5\% & 2.24 & 2.08 & 2.63 & 1.82 & 2.60 & 2.27 & 37.53\\
            EVQAScore& & 25\% & \textbf{2.79} & 2.88 & \textbf{3.19} & \textbf{2.22} & \textbf{2.92} & \textbf{2.80} & \textbf{39.50}\\
            EVQAScore& & 12.5\% & 2.66 & \textbf{3.03} & 3.08 & 2.05 & 2.60 & 2.68 & 38.69\\
            \bottomrule
        \end{tabular}
    }
    %\vspace{-4mm}
    \label{tab:VCG-Bench}
\end{table*}
This strong performance highlights EVQAScore's capacity to enhance existing scoring systems. It demonstrates that EVQAScore excels in evaluating video-caption data evaluation. 
%This further underscores the potential of EVQAScore for high-fidelity video-caption data assessment. As our method utilizes only keywords, previous caption-based methods suffer from the token limitation (Over 30\% of data is longer than the limitation of 77 tokens) of CLIP, making our approach more suitable for evaluating very long videos and their comprehensive caption. 
% Furthermore, PAC-S++ demonstrates a significant advancement over its predecessor PAC-S, achieving a Kendall $\tau_b$ score of 28.1 and a Spearman $\rho$ score of 36.4. When the PAC-S++ CLIP model is integrated with our EVQAScore framework, we observe an even greater boost, with Kendall $\tau_b$ and Spearman $\rho$ values of 32.8 and 42.3, respectively. This corresponds to a substantial improvement of 4.7 points in Kendall and 5.9 points in Spearman, showcasing the superior effectiveness of our approach when applied to an already enhanced baseline. 
% This strong performance highlights EVQAScore's capacity to amplify existing scoring systems, further emphasizing its potential for high-fidelity VideoQA data evaluation.

\subsection{EVQAScore for VideoQA Data Filtering}
% \begin{table}
%     \centering
%     \caption{Performance of VideoLLaVA and VideoLLaMA2 on MVBench using filtered data by PAC-S and EVQAScore.}
%     \resizebox{0.5\textwidth}{!}{ % 将表格等比例缩放到页面宽度
%         {\scriptsize % 缩小表格内容的整体字体
%         \begin{tabular}{lccc}
%             \toprule
%             \textbf{Method} & \textbf{Model} & \textbf{Data Proportion} & \textbf{Avg.} \\
%             \midrule
%             Baseline & \multirow{5}{*}{\textbf{VideoLLaMA2}} & 100\% & 34.09 \\
%             PAC-S 25\% &  & 25\% & 31.43 \\
%             PAC-S 12.5\% &  & 12.5\% & 34.24 \\
%             EVQAScore 25\% &  & 25\% & 33.34 \\
%             EVQAScore 12.5\% &  & 12.5\% & \textbf{36.65} \\
%             \midrule
%             Baseline & \multirow{5}{*}{\textbf{VideoLLaVA}} & 100\% & 30.68 \\
%             PAC-S 25\% &  & 25\% & 37.10 \\
%             PAC-S 12.5\% &  & 12.5\% & 37.53 \\
%             EVQAScore 25\% &  & 25\% & \textbf{39.50}  \\
%             EVQAScore 12.5\% &  & 12.5\% & 38.69 \\
%             \bottomrule
%         \end{tabular}
%         }
%     }
%     \label{tab:mvbench}
%     \vspace{-4mm}
% \end{table}

To address \textbf{Q2}, we perform data filtering using the EVQAScore metric in conjunction with the previous SOTA data evaluation baseline, PAC-S. The results presented in Table \ref{tab:data_filter} demonstrate the effectiveness of the EVQAScore metric in filtering VideoQA data. For both the top 25\% and 12.5\% data extractions, EVQAScore significantly reduces noisy data while retaining a larger proportion of high-quality VideoChatGPT and VideoChatGPT+ data compared to the PAC-S method. Specifically, EVQAScore retains only 456 and 15 noisy data points for the 25\% and 12.5\% extractions, respectively, while preserving a higher volume of high-quality data. This indicates that EVQAScore is more selective, thereby enhancing both data quality and processing efficiency for VideoQA tasks.

By reducing the amount of low-quality data, we enable the more efficient training of higher-performing VideoLLMs. Further experiments, presented in Section \ref{sec:videollm_performance}, demonstrate the impact of EVQAScore on VideoLLM performance.

\subsection{EVQAScore can Filter High-Quality VideoQA for VideoLLMs  Training}\label{sec:videollm_performance}
To address \textbf{Q3}, we train two models with different structures VideoLLaVA and VideoLLaMA2 on our filtered datasets. Then we conduct a comprehensive evaluation on several benchmarks, i.e. Activitynet, MSVD, MSRVTT, TGIF, MVBench, VCGBench and VCGBench Diverse. Our EVQAScore achieves SOTA performance across diverse models and benchmarks.
\paragraph{SOTA Performance Across Different Benchmarks}
The summarized results for ActivityNet, MSVD, MSRVTT, and TGIF are presented in Table \ref{tab:VCG-Bench}. It is evident that our EVQAScore achieves state-of-the-art (SOTA) performance on all benchmarks for both VideoLLaVA and VideoLLaMA2 models. Notably, EVQAScore demonstrates the ability to enhance long-video performance, as seen with ActivityNet (average video duration of 113.74s), as well as short-video performance, exemplified by TGIF (average duration of 11.93s). Furthermore, we extended the evaluation by utilizing our trained model on three more comprehensive benchmarks. The results for VCGBench are summarized in Table \ref{tab:VCG-Bench}, for VCGBench-Diverse in Table \ref{tab:vcgbench_diverse}, and for MVBench in Table \ref{tab:VCG-Bench}. Across all these datasets, our method consistently outperforms previous approach PAC-S. Moreover, our EVQAScore outperforms the 100\% data baseline in all benchmarks, this shows the importance of evaluating data quality and low quality data filtering.
\paragraph{SOTA Performance Across Different Models}
\begin{figure*}[ht]
  \includegraphics[width=\textwidth]{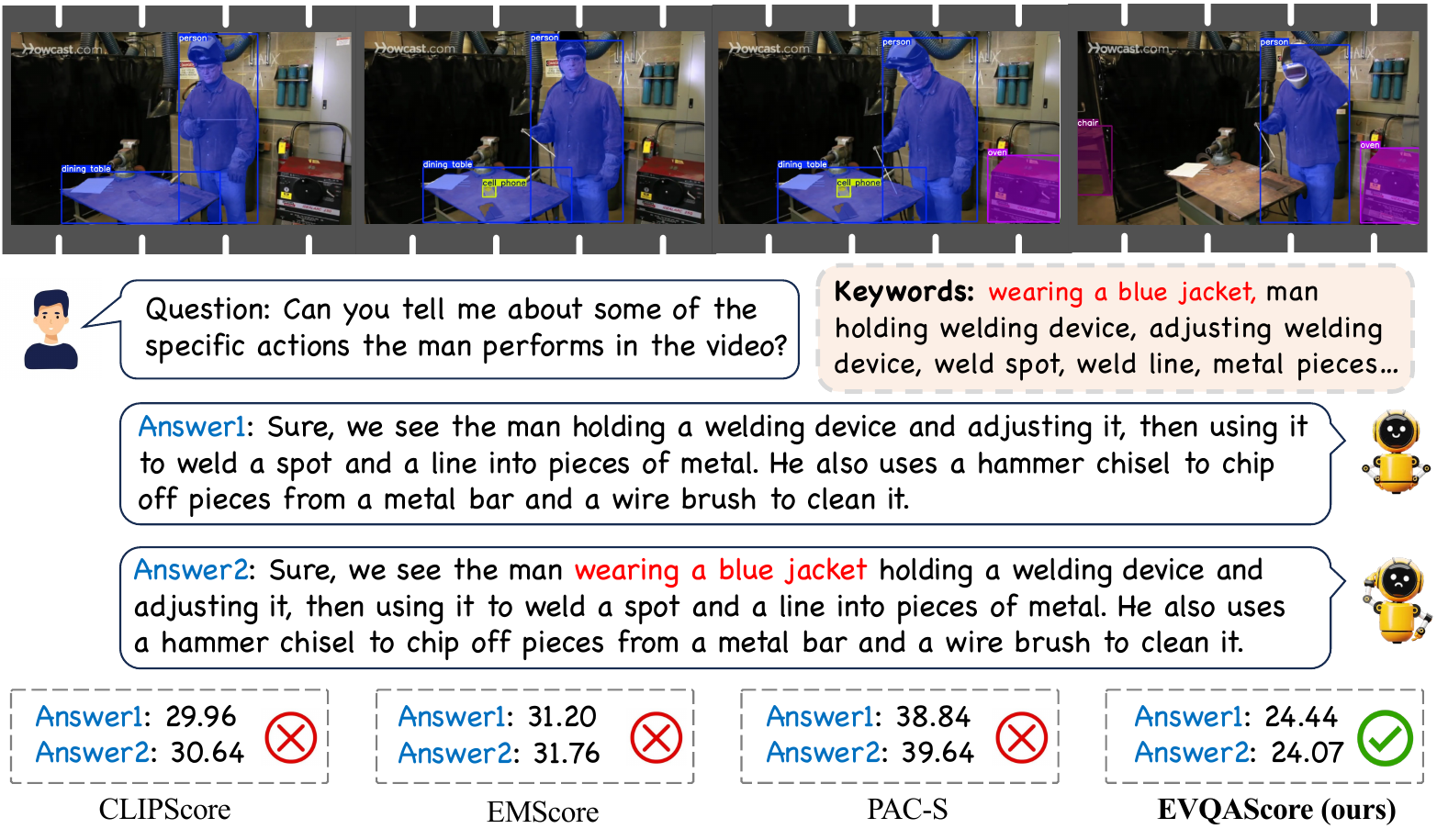}
  \caption{In this case study, Answer1 is more accurate, while Answer2 introduces hallucination. Our keyword extraction method successfully identifies hallucinations, making EVQAScore the only metric capable of determining that Answer1 is superior to Answer2.}
  \label{fig:Case_Study}
  %\vspace{-4mm}
\end{figure*}
We selected two different models: VideoLLaVA and VideoLLaMA2. These models have distinct architectures and use different encoders and foundation models. Specifically, VideoLLaVA employs LanguageBind as its encoder and Vicuna 1.5 as the foundation model, while VideoLLaMA2 utilizes CLIP as its encoder and LLaMA2 as the foundation model. Across the four benchmarks—ActivityNet, MSVD, MSRVTT, and TGIF presented in Table \ref{tab:Activitynet}, VCGBench and MVBench in Table \ref{tab:VCG-Bench}, VCGBench-Diverse in Table \ref{tab:vcgbench_diverse}—our filtered data outperformed the previous SOTA baseline, PAC-S, for both model architectures. This demonstrates the robustness of our EVQAScore.

Overall, we demonstrate that our EVQAScore effectively filters high-quality data to enhance VideoLLM performance. We show that EVQAScore not only evaluates video-caption data quality effectively, as detailed in section~\ref{sec:video-caption-evaluation}, but also improves the quality of VideoQA data and, consequently, the performance of VideoLLMs across various architectures and benchmarks. This underscores the broad applicability of our EVQAScore.

\subsection{Case Study: Evaluating VideoQA Data}
To address \textbf{Q5}, we present a Case Study to demonstrate the effectiveness of EVQAScore. As shown in Figure \ref{fig:Case_Study}, the user queries about the action depicted in the video. Answer1 provides a more accurate response, while Answer2 introduces hallucinations. Notably, our keyword extraction method successfully identifies the hallucinated part of Answer2, helping to distinguish it from the correct answer. This highlights the ability of EVQAScore to not only evaluate the relevance and accuracy of the answer but also to detect and penalize hallucinations. Therefore, EVQAScore correctly assigns a higher score to Answer1, reaffirming its superiority over Answer2.

\section{Conclusion}
In this work, we present EVQAScore, a reference-free evaluation method designed to assess the quality of both video-caption and VideoQA data for training VideoLLMs. By integrating patch retrieval, keyword extraction, and frame sampling, EVQAScore provides a fine-grained, efficient approach to video data quality evaluation. Our method surpasses existing techniques, achieving SOTA performance on the VATEX-EVAL benchmark, with significant improvements in Kendall and Spearman correlation scores. Additionally, EVQAScore enables effective data selection, reducing the data volume by over 87.5\% while maintaining high performance. The promising results demonstrate the potential of EVQAScore in enhancing the evaluation and selection of high-quality video data, contributing to the development of more efficient and accurate VideoLLMs. Our future work will explore further optimizations and broader applications of this method in other video-based tasks.

% \section{Limitations}
% Due to the limited computational resources, we didn't conduct experiments using all the videoQA datasets and all the VideoLLMs. Moreover, due to the limited funding, we use LLaMA for evaluation instead of GPT.
\clearpage
{
    \small
    \bibliographystyle{ieeenat_fullname}
    \bibliography{main}
}

% \appendix
% \clearpage
%\section{Keywords Extraction Prompt}

% \newpage
% {
%     \small
%     \bibliographystyle{ieeenat_fullname}
%     \bibliography{main}
% }

% WARNING: do not forget to delete the supplementary pages from your submission 
% \input{sec/X_suppl}

\end{document}